\documentclass[10pt,twocolumn,letterpaper]{article}

\usepackage[review,applications]{wacv}      

\usepackage{graphicx}
\usepackage{amsmath}
\usepackage{amssymb}
\usepackage{booktabs}

\usepackage[pagebackref,breaklinks,colorlinks]{hyperref}

\usepackage[capitalize]{cleveref}
\crefname{section}{Sec.}{Secs.}
\Crefname{section}{Section}{Sections}
\Crefname{table}{Table}{Tables}
\crefname{table}{Tab.}{Tabs.}

\def\wacvPaperID{1054} 
\def\confName{WACV}
\def\confYear{2025}

\begin{document}

\title{Transferring Foundation Models for Generalizable Robotic Manipulation}

\author{
Jiange Yang$^{1}$, Wenhui Tan$^{2}$, Chuhao Jin$^{2}$, Keling Yao$^{3}$, Bei Liu$^{4}$, \\
Jianlong Fu$^{4}$, Ruihua Song$^{2}$, Gangshan Wu$^{1}$, Limin Wang$^{1,5}$\thanks{Corresponding author.} \\
$^{1}$State Key Laboratory for Novel Software Technology, Nanjing University, China \\
$^{2}$Renmin University of China \\
$^{3}$The Chinese University of Hong Kong, Shenzhen \\
$^{4}$Microsoft Research \ \ \
$^{5}$Shanghai AI Lab\\
\tt\small jiangeyang.jgy@gmail.com, \{tanwenhui404, jinchuhao, rsong\}@ruc.edu.cn, \\
\tt\small 120090220@link.cuhk.edu.cn, \{jianf, Bei.Liu\}@microsoft.com, \{gswu, lmwang\}@nju.edu.cn
}


\maketitle

\begin{abstract}
  Improving the generalization capabilities of general-purpose robotic manipulation in real world has long been a significant challenge. Existing approaches often rely on collecting large-scale robotic data which is costly and time-consuming. However, due to insufficient diversity of data, they typically suffer from limiting their capability in open-domain scenarios with new objects and diverse environments. In this paper, we propose a novel paradigm that effectively leverages language-reasoning segmentation mask generated by internet-scale foundation models, to condition robot manipulation tasks. By integrating the mask modality, which incorporates semantic, geometric, and temporal correlation priors derived from vision foundation models, into the end-to-end policy model, our approach can effectively and robustly perceive object pose and enable sample-efficient generalization learning, including new object instances, semantic categories, and unseen backgrounds. We first introduce a series of foundation models to ground natural language demands across multiple tasks. Secondly, we develop a two-stream 2D policy model based on imitation learning, which processes raw images and object masks to predict robot actions with a local-global perception manner. Extensive real-world experiments conducted on a Franka Emika robot and a low-cost dual-arm robot demonstrate the effectiveness of our proposed paradigm and policy. Demos can be found in \href{https://www.youtube.com/watch?v=MAcUPFBfRIw}{\color{red}link\underline{1}} or \href{https://www.youtube.com/watch?v=1m9wNzfp_4E&t=1s}{\color{blue}link\underline{2}} and our code will be released at https://github.com/MCG-NJU/TPM.
\end{abstract}


\begin{figure*}[h]
    \centering
    \includegraphics[width=0.95\linewidth]{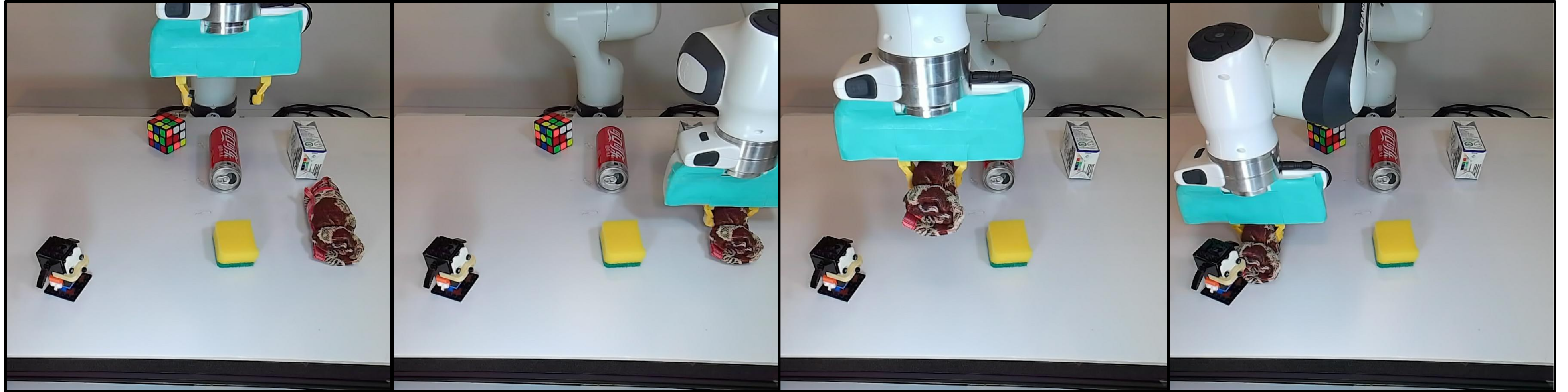}
    \caption{A demonstration of our task. Receiving human instruction ``I want to take a shower'', our model can reason out the desired object (i.e., the towel), and then precisely pick and place it near the target object (i.e., the user represented by a Lego toy).}
    \label{fig:demo}
\vspace{-3mm}
\end{figure*}


\section{Introduction}
\label{sec:intro}


Creating a general-purpose robotic agent that is capable of performing diverse tasks in real-world remains a long-standing and challenging research. In this paper, we endeavor to develop a model that enables generalizable robotic manipulation. The first challenge in creating generalizable agents is effectively converting abstract task condition instructions into specific robot inputs. Current approaches utilize various forms, including task identifiers~\cite{identifiers}, goal images~\cite{goal}, video showcasing of human demonstrations~\cite{bcz}, and natural language~\cite{rt1,bcz,language,cliport,hiveformer}. Language, in particular, provides a natural and scalable manner for human-robot interaction, but may be under-specified and ambiguous. The second challenge involves enhancing the generalization capabilities of a single robot model to handle multiple tasks, encompassing both unseen objects and environments. To address the above challenges, recent advancements~\cite{bcz,rt1,saycan} have predominantly embraced data-driven learning-based methods. Notably, one of the groundbreaking work is RT-1~\cite{rt1}, which has introduced a comprehensive model capable of executing diverse instructions using an extensive dataset of approximately 130,000 demonstrations spanning over 17 months and involving 13 robots. However, the collection of real-world data poses significant resource requirements, and the approach exhibits limitation to compositional generalization, struggling with unseen objects and environments, due to insufficient diversity of data.

In this paper, we propose to achieve sample-efficient generalization for robotic manipulation by introducing language-reasoning mask modality containing semantics, geometry, and temporal correlation priors inherent from internet-scale vision foundation models into an end-to-end policy model. Specifically, segmentation has been proven significant as grasping priors in manipulation tasks~\cite{grasps,see,grasping}, and we introduce language-reasoning mask as a new condition modality for policy model and conduct end-to-end training using imitation learning. An intuitive pipeline for robot manipulation is to first achieve visual perception and then conduct motion planning like ~\cite{codex,progprompt,twostage}. However, such pipeline necessitates an efficient motion planner, while also requiring completely accurate object mask and depth for constructing point clouds, which may not perform well when dealing with transparent and disturbed objects, as well as unstructured environments with collision situation. In contrast, the end-to-end 2D policy pipeline we adopt in this work, as well as other works~\cite{bcz,rt1}, can dynamically receive raw image as input and output continuous action in a close-loop manner, which does not rely on depth calibration and completely accurate object masks. Furthermore, \textit{our paradigm aims to fully unify the generalizability of internet-scale models and the potential of imitation learning to capture multimodal action distribution from human skills at the lowest possible training and inference cost, while also mitigating the ambiguity of language as condition. Therefore, this paradigm is highly scalable.}

In order to further build a holistic robot system with human-robot interaction, we utilize large language model to reason human demands and design a two-stream policy model to predict actions with a local-global perception manner. Specifically, we first use GPT-4~\cite{gpt4} to interpret language instructions and generate desired object prompts. Second, we identify and locate objects by open-vocabulary detection~\cite{dino} and tracking~\cite{mixformer} models. We then adopt the vision foundation model SAM~\cite{sam} to generate segmentation masks of desired objects. Subsequently, we propose a \textbf{T}wo-stream architecture \textbf{P}olicy \textbf{M}odel, \textbf{TPM}, which uses a deeper branch to capture global RGB information and a shallower branch to capture local object-related RGB-M information as well as fuses multi-view features and robot proprioception states through attention mechanism. These designs enable more robust 3D perception and thus leading to accurate action prediction. Finally, to verify our system, we primarily select the widely popular pick-and-place (picking A and placing near B) tasks for quantitative evaluation. Specifically, we collect a dataset consisting of 1000 demonstrations with 40 objects for training. Our experiments results show the effectiveness of our proposed paradigm and policy model architecture, particularly in generalizing to unseen objects, complex backgrounds, and multiple distractors. A simple demonstration of our task is shown in Figure~\ref{fig:demo}. In addition, we conduct further experiments to show the capabilities of our method to transfer more manipulation skills, including opening drawer, picking A and placing inside of B, placing on top of B, folding and stacking.

In summary, the contributions of our paper are summarized as follows:
\begin{itemize} 
\item We propose a novel paradigm by transferring internet-scale foundation models for robotic manipulation with language-reasoning segmentation mask, aiming to enhance its generalization capabilities in a sample-efficient way. Additionally, the mask modality also provide a more specified and unambiguous condition representation than unprocessed human language.

\item We develop a two-stream policy model for handling images and language-reasoning masks with a local-global perception manner, which achieves better spatial relationship understanding.

\item Extensive real-world experiment results demonstrate that our paradigm and policy model architecture can effectively improve the performance and generalize to handle unseen objects, new backgrounds, more distractors, and even expand to more manipulation skills.
\end{itemize}  

\section{Related Work}

\textbf{Pre-trained Foundation Models for Robotics.} Recently several works~\cite{saycan,zeroshot,Monologue,Grounded,text2motion,codex,progprompt,voxposer,msgpt,ok} explored to leverage off-the-shelf large language models to plan feasible tasks for robotics. Some works~\cite{palme,alphablock,embodiedgpt,detgpt,rt2,openx} further use in-domain data to fine-tune LLMs for embodied reasoning. InstructRL~\cite{instruction} employs masked autoencoder M3AE~\cite{M3AE} to encode visual observations and language instructions. PAFF~\cite{policy} utilizes CLIP~\cite{clip} to provide feedback for relabeling demonstrations. MOO~\cite{moo} leverages pre-trained vision-language model to extract object-centric representations, which is based on a single pixel in the center of the bounding box on the first frame. In contrast, our approach distinctively employs more specified object masks using a detection-tracking-segmentation manner, which not only enhances the precision and reliability of object representation for robotic manipulation tasks, but also provides stronger geometric and temporal correlation priors. In addition, different from~\cite{codex,progprompt,voxposer}, which use prompt-based segmentation mask and depth to construct object point cloud and then call for additional grasp and motion planner, \textit{we directly utilize the mask conditioning the end-to-end and learnable policy model with closed-loop manner, without requring depth, camera calibration and completely accurate object masks}.

\textbf{Vision-based Robot Learning.}
Vision-based robot learning plays a crucial role in robotics. Recently pre-trained visual models for robotics have been rapidly developing~\cite{r3m,mvp,vip,liv,robotclip,driven,vc1,stp,spa}. R3M~\cite{r3m} and VC-1 \cite{vc1} demonstrate the effectiveness of pre-training on egocentric videos. STP~\cite{stp} further considers temporal motion cues. \cite{vip} proposes a vision model capable of producing dense reward
signals and LIV \cite{liv} expand it to multi-modal. Li et al. show that pre-training on semantic tasks like classification and segmentation helps in improving efficiency and generalization of grasping~\cite{see}. Some works~\cite{videodex,mimicplay,plan,atm,tramoe,flow,gen2act} also focus on learning skill priors from large-scale human video data. As for model architecture, several works~\cite{explore,vision} point out that convolution-based models tend to outperform transformer-based models and ~\cite{diffusion,act} demonstrates the superiority of generative modeling of the policy. Our work is orthogonal to the contributions of these works. In addition, some works~\cite{gpd,dexnet,contactgraspnet,anygrasp} focus on point cloud-based grasping pose generation. In contrast, our method requires dynamically mapping observations to continuous actions with closed-loop manner. With different problem formulations, direct comparisons are usually not performed.


\textbf{Language-Conditioned Robotics Control.} The goal of building a robotic model that can follow diverse natural language instructions has consistently been an active research field~\cite{rt1,bcz, language,cliport,hiveformer,perceiver, affordances,matters,calvin,rvt,act3d}. Noteworthy advancements have been made by various approaches. Hiveformer~\cite{hiveformer} proposes a unifed approach to encode the full history of observation-action pairs. Perceive-actor~\cite{perceiver} encodes RGB-D voxel observations with a Perceiver Transformer to provide a strong structural prior. BC-Z~\cite{bcz} and RT-1~\cite{rt1} focus on scaling and expanding the collection of real-world data to facilitate generalization of robots, which encompasses unseen tasks, environments, and objects. However, due to the limitations in both quantity and diversity of the collected data compared to large-scale datasets in the vision and language domains~\cite{imagenet,laion,hdvila}, the generalization ability of the trained models is still relatively poor, especially in unseen object categories.

\vspace{-2mm}

\section{Approach}
In this section, we begin by presenting the problem formulation of our method in Section~\ref{subsec:formulation}.
Following this, we detail the pipeline of language-reasoning mask generation with foundation models in Sections~\ref{Workflow}.
Finally, we elaborate on our two-stream policy model and its training methodology in Section~\ref{policy2}. We primarily illustrate our approach using pick-and-place skill, and the same manner can be applied to other skills as well.

\begin{figure*}[htbp]
\centering
\includegraphics[width=0.98\textwidth]{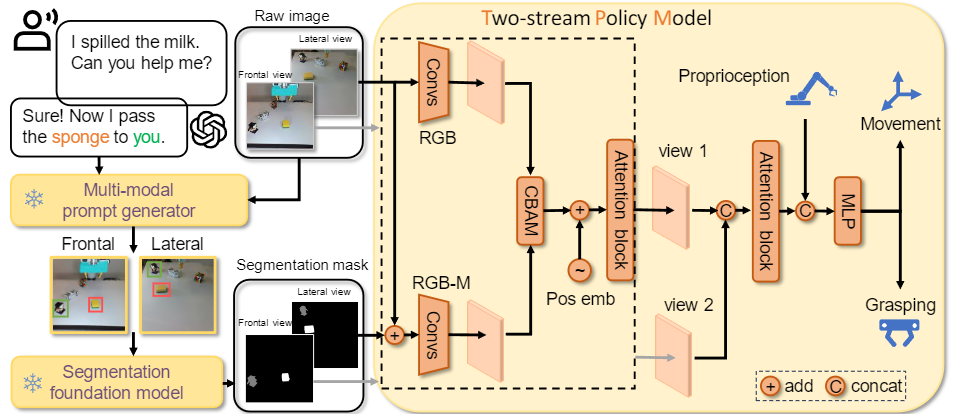}
\caption{
Our model comprises four components:
(1) GPT-4 reasons target objects based on human demands. (2) A multi-modal prompt generator, comprising object detection and tracking models, transforming input images and target object prompts into bounding boxes. (3) The Segment Anything model, which uses bounding boxes as prompts to segment target objects and generate task-relevant masks. (4) A two-stream policy model that processes images, language-reasonin segmentation masks, and robot proprioception to predict actions.}
\vspace{-5mm}
\label{fig:model_architecture}
\end{figure*}


\subsection{Problem formulation}\label{subsec:formulation}
We aim to develop a generalizable robot capable of interpreting high-level language instructions from users and executing precise actions to fulfill their needs. For instance, when a user says, ``I am thirsty'', the robot should pick the milks up from the tabletop, and place it near the user's hands. This process seamlessly integrates perception, reasoning, and control into a unified pipeline.

In our problem setting, we aim to develop a robotic system, $\Phi_{\varepsilon}(a|p,o,l)$, parameterized by $\varepsilon$, which maps robot proprioception $p$, visual observations $o$, and human-provided language instructions $l$ to continuous actions $a$ on a physical robot.
To enhance the generalization capabilities and sample efficiency, we divide our system into two parts:
(1) A segmentation mask generator based on foundation models discussed in Section~\ref{Workflow}, which fully leverages the potential of internet-based foundation models to obtain the desired object masks based on human instructions.
(2) A two-stream policy model described in Section~\ref{policy2}, which encodes multi-modal inputs, including raw images, language-reasoning object masks, and robot proprioception, mapping them to specific robot actions.

\subsection{Mask Generation with Foundation Models}\label{Workflow}

To enhance the generalization capabilities of robot agents, we propose a novel paradigm that utilizes a series of internet-scale foundation models to accurately interpret abstract natural language instructions, locate objects, and segment their geometry mask for subsequent specified task condition. Specifically, this paradigm comprises target object reasoning based on LLMs, a multi-modal prompt generator based on open-vocabulary detector and tracker, and segmentation mask generation based SAM. The framework of our robotic agent system is provided in Figure~\ref{fig:model_architecture}.

\textbf{Target objects reasoning.} We employ an LLM, GPT-4~\cite{gpt4}, as the central cognitive component of our model. When a user submits a high-level language instruction (e.g., ``I spilled the milk''), it is integrated into a designed prompt template, which is subsequently fed into the GPT-4 to reason out the target objects for task condition (e.g., pick a sponge and place it near the user). This process enables our model to effectively deduce the target objects that fulfills the user's requirements, ensuring the correct task condition. 

\textbf{Multi-modal prompt generator.} After reasoning out which targets should be interacted, we apply a state-of-the-art open-vocabulary object detector Grounding DINO~\cite{dino} with powerful semantic concept generalization ability to locate the target objects given language expression. However, during the process of a robotic agent performing a task, the target to be manipulated inevitably encounters occlusion (e.g., being blocked by the robotic arm), disturbance (e.g., dynamic objects) and potential distractors. Compared to object detection, object tracking is more robust in handling these challenging scenarios due to its inherent spatio-temporal correlation. Therefore, after the robotic agent completes the first step of the action, we switch to a state-of-the-art and efficient tracker MixFormer~\cite{mixformer} to obtain all subsequent bounding boxes. To conclude, we obtain the bounding box by a object detector in the first step, and tracks it after this, which consequently serve as prompts for object mask generation.

\textbf{Mask generation based on SAM.}
The image segmentation model SAM~\cite{sam} has been widely explored in various fields due to its powerful object generalization capability and promptability. The appeal of SAM is enhanced by its ability to flexibly integrate the semantic concept generalization of open-vocabulary detectors with a bounding box prompt. In robot manipulation tasks, object masks are closely related to affordance maps because of their shape and geometry prior. Thus, after identifying and locating the target objects, we provide the bounding box as a prompt to SAM to generate the segmentation masks of them.

In our approach, we transform abstract language instructions into specified target object masks, which incorporates rich semantic, geometry, and temporal correlation priors derived from vision foundation models into the end-to-end 2D policy model. By collecting a small amount of real robotic data, the policy model is able to efficiently learn how to adapt these capabilities onto specific manipulation actions, thereby achieving sample-efficient learning of generalizable manipulation.
In summary, the process of target object masks generation can be formulated as follow:
\begin{equation} 
m = \mathrm{SAM}(o,\mathrm{PG}(o,\mathrm{RM}(o,l)) ),
\end{equation}
where PG and RM denote the multi-modal prompt generator and the reasoning model, respectively. $m$, $o$ and $l$ stand for the generated object masks, raw images and language instructions, respectively.

\subsection{Two-stream Policy Model Architecture}\label{policy2}

\textbf{Policy model architecture.}
As shown in the right side of Figure \ref{fig:model_architecture}, the two-stream Policy Model, which we refer to as \textbf{TPM}, maps the robot proprioception, object masks, as well as the raw image observations of two different views to continuous actions. The process can be formulated as
\begin{equation}
    \pi_{\theta}(p, (o_{1},m_{1}),(o_{2},m_{2})),
\end{equation}
where the subscripts index different perspective of views, and TPM is parameterized by $\theta$.

Inspired by the success of mask features for memory bank in video object segmentation~\cite{stcn}, we concatenate the image RGB frame and object mask along the channel dimension to form RGB-M. To make it clear, we further explain details as follows: (1) we first adopt a two-stream convolution-based architecture to separately encode RGB-M and RGB, where RGB-M branch employs a shallower ResNet-18~\cite{resnet} network to capture local features of task-related objects, and RGB-only branch utilizes a deeper ResNet-50 network to capture the global spatial relationships within the entire scene. We both take stage-4 features with stride 16 as standard feature maps, and utilize a CBAM~\cite{CBAM} block to fuse them from space and channel dimension.
(2) In order to enhance the spatial perception capability, we add a 2D positional embedding for each feature point and subsequently apply an attention~\cite{transformer} block to promote spatial interaction. As for multi-view feature fusion, we concatenate two view features and also employ a global self-attention block to ensure dynamic spatial alignment. (3) Finally, for better integrating embodied perception to our policy model, we further inject the robot proprioception (pose of the end-effector) state. In our work, the proprioception embeddings and visual embeddings are fused via concatenation to obtain the final state representation, which is then fed into two MLPs, to generate the predicted next action, i.e., the movement along x, y and z-axis and the opening state of the gripper.

\textbf{Policy model training.}
We train our TPM using behavior cloning with our collected offline dataset $\boldsymbol{\mathcal{D}_1}$. In overall, the loss function can be formulated as follow:
\begin{equation} 
\min_{\theta} {\textstyle \sum_{  \boldsymbol{\mathcal{D}}}} \ \mathrm{CE}(a_g,\pi_{\theta} (p, o, m)) + \mu \cdot \mathrm{MSE}(a_m,\pi_{\theta} (p, o, m)),
\end{equation}
where $a_g$ and $a_m$ denote the gripper state and the movement of the robot's end-effector. We adopt a deterministic policy manner. The model learns the gripper open or closed state as a binary classification task with cross-entropy loss (CE), and learns the continuous movement of end-effector through mean square error (MSE). The $\mu$ denotes the loss weight of the MSE loss.

\vspace{-2mm}

\section{Experiment}
In this section, we first introduce our pick-and-place dataset. Then we elaborate on the implementation details, experimental setup and evaluation 
 results in the following sections. Finally, we elucidate how our pipeline can be flexibly extended to other skills, and provide both qualitative and quantitative experiment results.

 \vspace{-3mm}

\subsection{Dataset}
\begin{figure*}[t]
    \centering
    \includegraphics[width=0.97\textwidth]{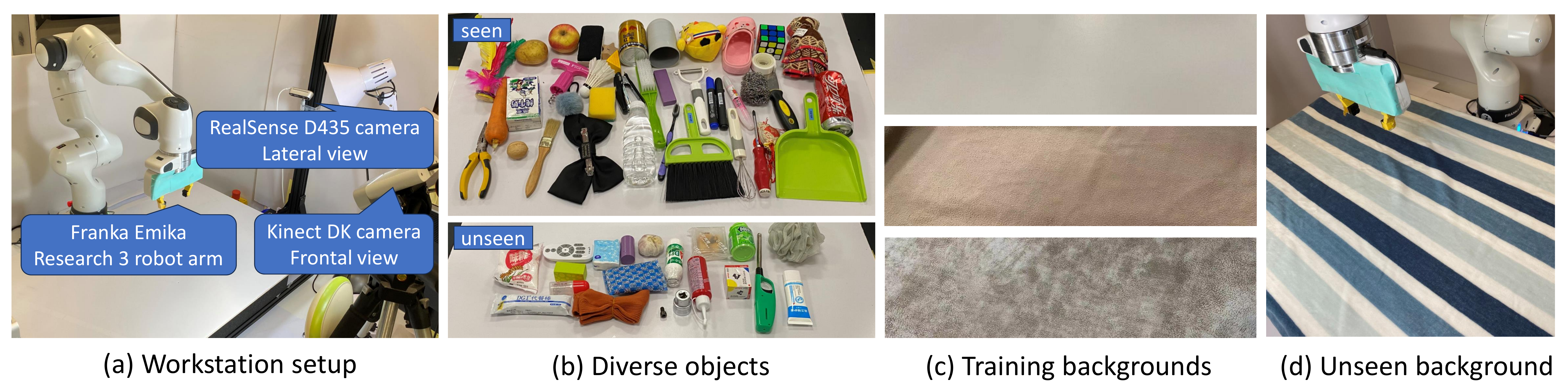}
    \caption{(a): Overview of our workstation, which has a Franka robot arm, a frontal view camera, and a lateral view camera. (b): Seen and unseen objects in the experiments. (c): Three backgrounds in the training data. (d): A challenging background with complex texture for new background evaluation.}
    \label{fig:workstation}
\end{figure*}

To facilitate the experiments of our work, we collect a dataset using Franka Emika Research 3 robot arm to perform imitation learning, which contains 1000 episodes. For each episode, we annotate the language instruction and use GroundingDINO~\cite{dino} and SAM~\cite{sam} to obtain the language-conditioned masks. We select manipulated objects of various types, such as those with different shapes, sizes, textures, and colors, to ensure the diversity of the dataset. Specifically, we capture multi-view images from both frontal and lateral perspectives. We select 40 common objects categorized into 5 typical shapes from daily life, in a total of 1000 pick-and-place trajectories across 3 different table-top backgrounds, as shown in Figure~\ref{fig:workstation}(c). For every pick-and-place demonstration, we randomly place 0 to 2 additional objects as distractors.

\subsection{{Implementation} Details}

{In our evaluation experiments, all vision foundation models~\cite{sam,dino,mixformer} employ the base version~\cite{vit,swin}.} To optimize our proposed TPM, we use Adam~\cite{adam} with decoupled weight decay~\cite{weight} of $5{\cdot}10^{-4}$. The peak learning rate is set to $5{\cdot}10^{-5}$, decaying with a cosine learning rate schedule to $5{\cdot}10^{-6}$. We empirically set the weight $\mu$ of movement loss to gripper state loss at 1,000 to keep them in comparable magnitude. We train on 224$\times$224 images without data augmentation, with a batch size of {24} for {500k} iterations. The model code is implemented in PyTorch~\cite{torch} and trained on an NVIDIA RTX A6000 GPU.

\subsection{Experimental Setup}\label{sec:setup}

\textbf{Real-World Environment.} In our real-world experiments, we use a Franka Emika Research 3 robot arm in a table-top environment, consistent with data collection. The Intel RealSense camera (1,280$\times$720 resolution) and Azure Kinect DK camera (2,048$\times$1,536 resolution) are mounted on fixed supports, as shown in Figure~\ref{fig:workstation}(a). 

\textbf{Evaluation Metric.} Our quantitative experiments primarily focus on pick-and-place tasks described as ``pick A and place it near B''. We measure the percentage of successful pick-and-place tasks as success rate. A successful pick-and-place is defined as (1) grasping A and (2) placing it within 2 inches of B.

\textbf{Evaluation Setup.}
We define the {40} objects in the collected training data as seen objects and hold out another 20 objects not present in the training data as unseen objects. To comprehensively evaluate the generalization capabilities of our method, we evaluate our model from three aspects:
\textit{(1) Seen/Unseen objects}.
We have two settings, one is with seen objects for both A and B and the other is with unseen objects for A and B. Unlike prior works~\cite{rt1} focusing on compositional generalization for unseen tasks, our unseen objects are strictly new {categories or instances}, testing the model's ability to handle new objects. We present some objects in Figure~\ref{fig:workstation}(b).
\textit{(2) New background}. We introduce an environment with a complex-textured tablecloth, altering lighting, materials, and backgrounds, at the same time. This evaluates our model's robustness to out-of-distribution generalization. We show the new background in Figure~\ref{fig:workstation}(d).
\textit{(3) More distractors}. We add a scenario with more than 2 distractor objects (ranging from 3 to 6 objects), creating a congested tabletop scene with target objects disturbance and occlusion. This evaluates the model's robustness against more distractor objects.


\begin{table}[t]
\centering

\label{tab:main_exp}
\begin{tabular}{lccc}
\toprule
Scenario       & Seen      & Unseen & Average   \\ \hline
Standard                   & {82.5}  & {80.0} & {81.25} \\
New background & {65.0}  & {55.0} &  {60.0}\\
More distractors      & {75.0} & {70.0} & {72.5} \\ 
\bottomrule
\end{tabular}
\caption{Experimental results evaluated on different scenarios.}
\label{tab:main_exp}
\vspace{-6mm}
\end{table}

\subsection{Experiment results}

\begin{figure}[htbp]
    \centering
    \label{other}
\includegraphics[width=0.495\textwidth]{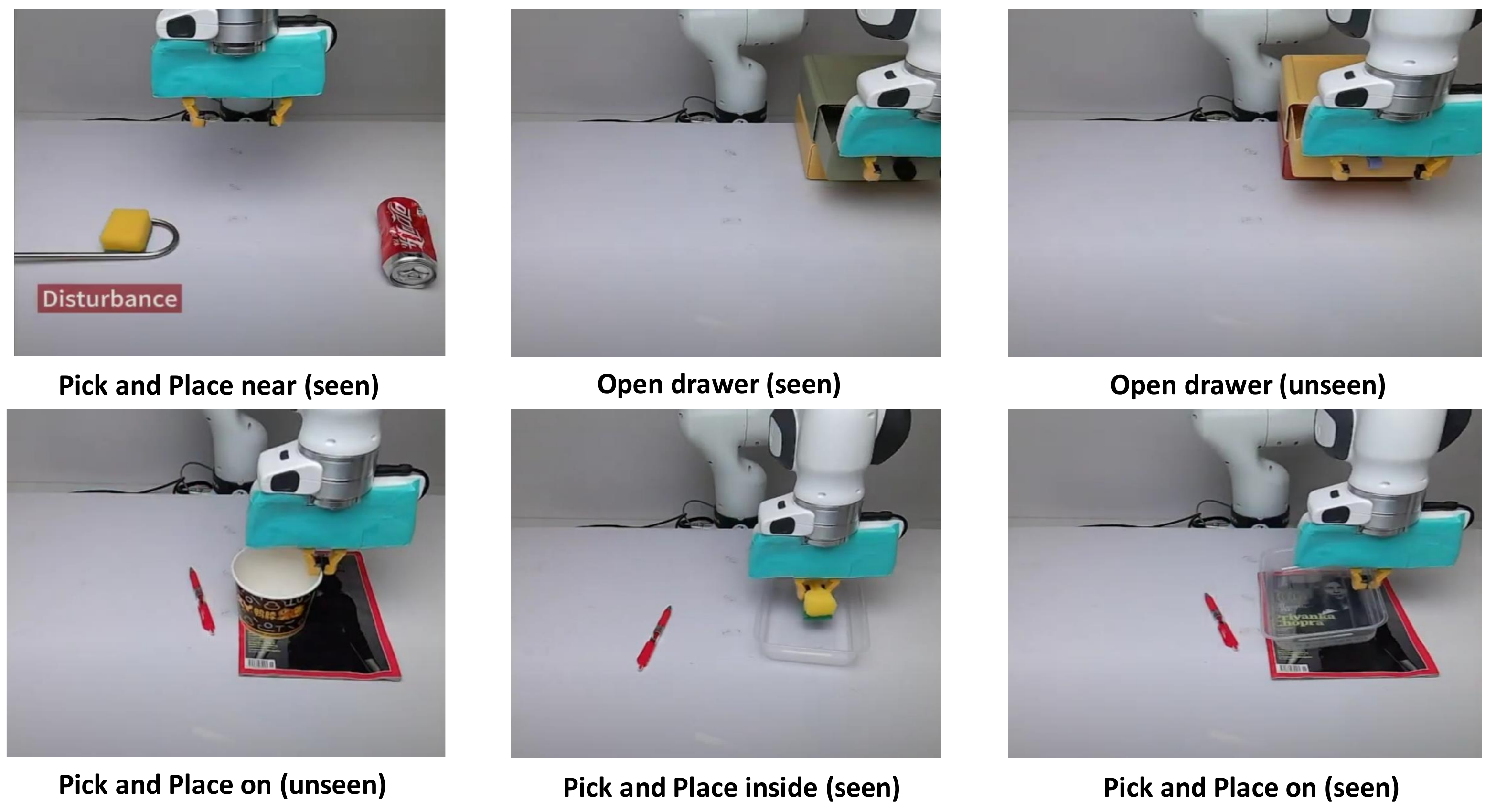}
    \caption{Some demonstration examples of disturbances scene and other manipulation skills.}
    \label{other}
\end{figure}

\begin{figure*}[htbp]
    \centering
\includegraphics[width=0.97\textwidth]{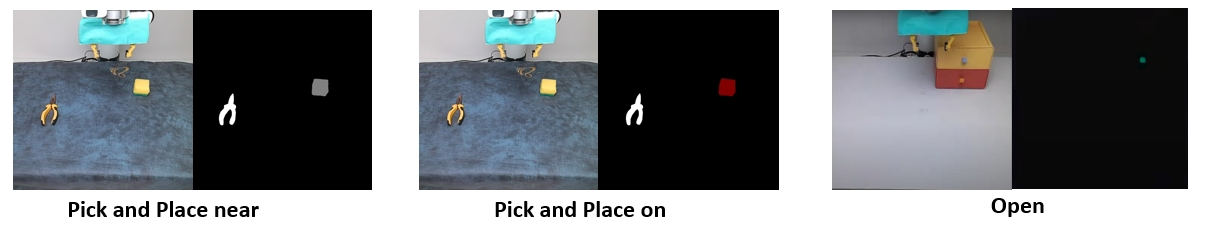}
\caption{Our policy model can be conditioned by assigning different values to object masks for different manipulation skills.}
    \label{seg}
\end{figure*}

We initially conduct a comprehensive evaluation of our method's effectiveness and its ability to generalize to unseen objects, new backgrounds, and more distractors. To ensure a more extensive evaluation, we randomly select 20 tasks under each setting. For each task, the source object initial positions are placed once on the left and once on the right, resulting in a total of \textbf{40} trials. The \textit{standard} environment refers to a tabletop background with 0-2 distractors, consistent with the training data.
The experiment results are presented in Table \ref{tab:main_exp}. These results indicate slight performance decrease when introducing a new background. However, our model demonstrates robustness to a greater number of distractors and unseen objects, which exceeds the limit of RT-1 and can be attributed to the inclusion of language-reasoning segmentation mask modality derived from foundation models for action prediction.

\begin{table*}[htbp]
\centering

{
\begin{tabular}{lccccc}
\toprule
Method            & \multicolumn{1}{l}{Seen} & \multicolumn{1}{l}{Unseen} & \multicolumn{1}{l}{New background} & \multicolumn{1}{l}{More distractors} & Average\\ \hline
Ours              & {82.5}                 & {80.0}                       & {65.0}                             & {75.0}   &     {\textbf{75.625}}                               \\
\quad -MOO-like~\cite{moo}          & {50.0}                     & {42.5}                       & {27.5}                             & {35.0}      &    {38.75} \\
\quad -RT-1-like~\cite{rt1}          & {65.0}                     & {0.0}                       & {20.0}                             & {60.0}      &    {36.25}  \\  \hline
\quad -replace mask with bbox          & {50.0}                     & {40.0}                       & {25.0}                             & {30.0}      &    {36.25}                             \\
\quad -w/o tracking & {70.0}                      & {50.0}                       & {55.0}                             & {70.0}      &     {61.25}                            \\
\quad -single view  & {65.0}                     & {80.0}                      & {20.0}                                & {70.0}      &       {58.75}                         \\
\quad -RGB-M only   & {85.0}                     & {70.0}                      & {50.0}                             & {70.0}        &       {68.75}                        \\ 
\bottomrule
\end{tabular}}
\caption{Comparison of our method and its variants on various settings.}
\vspace{-5mm}
\label{ablation}
\end{table*}

To further verify the effectiveness of our proposed approach and its individual components, we compare our method with several variants. We evaluate each method in four settings: (1) seen objects in the standard environment; (2) unseen objects in the standard environment; (3) seen objects in a new background; and (4) seen objects with more distractors (randamly 3-6). Each setting includes 20 tasks, resulting in a total of \textbf{80} trials for each method. The comparative experiment results are presented in Table~\ref{ablation}. Additionally, we also follow the setup in Table~\ref{tab:main_exp}, replacing the masks obtained from foundation models with initial language instructions, called as RT-1-like. The experiment results show that due to overfitting, this scheme is comprehensively inferior to our paradigm, especially in the setting with unseen objects. Finally, we analyze the results to address the following questions:
\begin{itemize}
\setlength{\itemsep}{0pt}
\setlength{\parsep}{0pt}
\setlength{\parskip}{0pt}
    \item Does the segmentation mask outperform the bounding box for action prediction?
    \item Is tracking more robust for prompt generation than frame-by-frame detection?
    \item Is attention based multi-view fusion more advantageous than a single front view?
    \item Does incorporating a separate RGB-only branch yield better performance?
\end{itemize}

\textbf{Segmentation mask is more effective than bounding box for action prediction.} A related work MOO~\cite{moo} uses object-centric pixel and necessitates the model to implicitly perform temporal correlation due to fixing the object-centric pixel after the first frame. To re-implement it, we attempt to extend our model to a temporal version by adding temporal attention and use the center pixel of the bounding box in the first frame to condition tasks. However, we find that this model even struggles with picking correct object instances. We conjecture that point-based prompt learning is relatively challenging and prone to over-fit in our limited low-data regime. Additionally, we also replace the object mask with its bounding box. The results (75.625 vs. 36.25) show that the segmentation mask significantly outperforms the bounding box. In addition to provide more rich geometry and shape priors, the segmentation mask also demonstrates greater robustness to complex textures and distractors. In contrast, the bounding box struggles to achieve such precision. Moreover, explicitly incorporating a tracker allows our model to easily handle dynamic objects or those subjected to disturbances, as shown in Figure~\ref{other}.

\textbf{Tracking is more robust for prompt generation than frame-by-frame detection.} We then replace the paradigm of first-frame detection and subsequent-frame tracking with frame-by-frame detection for prompt generation. The average success rate significantly decreases, particularly when the robot arm severely obstructs the object during grasping, illustrating the robustness of the detection-tracking paradigm. Moreover, the detection-tracking paradigm also substantially improves the inference speed.

\textbf{Multi-view fusion is more beneficial compared to single view.} We further investigate the conversion of the multi-view model to a single-view model, retaining only the front view. The experiment results  demonstrate the effectiveness of multi-view fusion. Specifically, there is a significant drop (65.0 vs. 20.0) in the new background. We believe this is because multi-view vision can estimate depth through disparity, making it more robust than single-view vision.

\textbf{Incorporating a separate RGB branch is beneficial.} Finally, we implement a single-branch architecture RGB-M policy model based on ResNet-50 for a fair comparison. The experiment results  demonstrate the effectiveness of our two-stream architecture. This two-stream approach allows the model to better capture both local and global features. Additionally, the disparity (50.0 vs. 65.0) between the two models is most noticeable when altering the background, with the RGB-M model even exhibiting hovering motion in mid-air. We hypothesize that incorporating an RGB-only branch could contribute to better generalization in depth estimation, and a separate RGB-M branch may be more prone to overfitting the distribution of the training data.

\begin{table*}[tbp]
\label{llm}
\centering

\begin{tabular}{cccc}
\toprule
             & GPT-4~\cite{gpt4} & DetGPT~\cite{detgpt} & MiniGPT-4~\cite{minigpt4} \\
Success Rate & \textbf{0.95}  & 0.75   & 0.2       \\ 
\bottomrule
\end{tabular}
\caption{The reasoning performance comparison of LLMs.}
\label{llm}
\end{table*}

\begin{table*}[htbp]  
\centering

\begin{tabular}{ccccccc}  
\toprule
& GroundingDINO-B & Mixforemr-B & MixformerV2-S & SAM-B & SAM-T & TPM \\
Inference Time (ms) & 148.6 & 103.4 & 17.0 & 18.2 & 10.1 & 34.8  \\ 
\bottomrule
\end{tabular}  
\caption{The inference time for different modules and model sizes.}
\vspace{-5mm}
\label{speed}
\end{table*}


\subsection{Extension to other skills}

{Although we mainly demonstrate the effectiveness of our paradigm and policy on quantitative pick-and-place tasks evaluation, our multi-task policy model can be conditioned by assigning different values to object masks for different manipulation skills, as shown in Figure~\ref{seg}. Therefore, our method can be flexibly extended to some common skills by transferring language instruction of skills to mask values of manipulated objects. To verify this assumption, we further collect 100 demonstrations of other tasks to co-fine-tune our model, including ``open drawer'', ``pick A and place inside of B'', and ``pick A and place on top of B''. Some  demonstration examples are shown in Figure~\ref{other}. For drawer-opening task, we mask the handle of the drawer based on the output of GPT-4, and conduct \textbf{40} trials, achieving a success rate of approximately \textbf{50\%}. More qualitative results of other skills can be found in the demo video. 

Additionally, we also validate folding and stacking skills on a low-cost dual-arm robot. Specifically, we co-fine-tune the previous TPM model weight using 50 new demonstrations and deploy it using Mixformer-small~\cite{mixformerv2} and SAM-tiny~\cite{mobile}. The demos can be seen Fig~\ref{222}. It is worth mentioning that our low-cost dual-arm does not require additional motion planning time due to without movelt calls.


\begin{figure*}[htbp]
\centering
\includegraphics[width=0.95\textwidth]{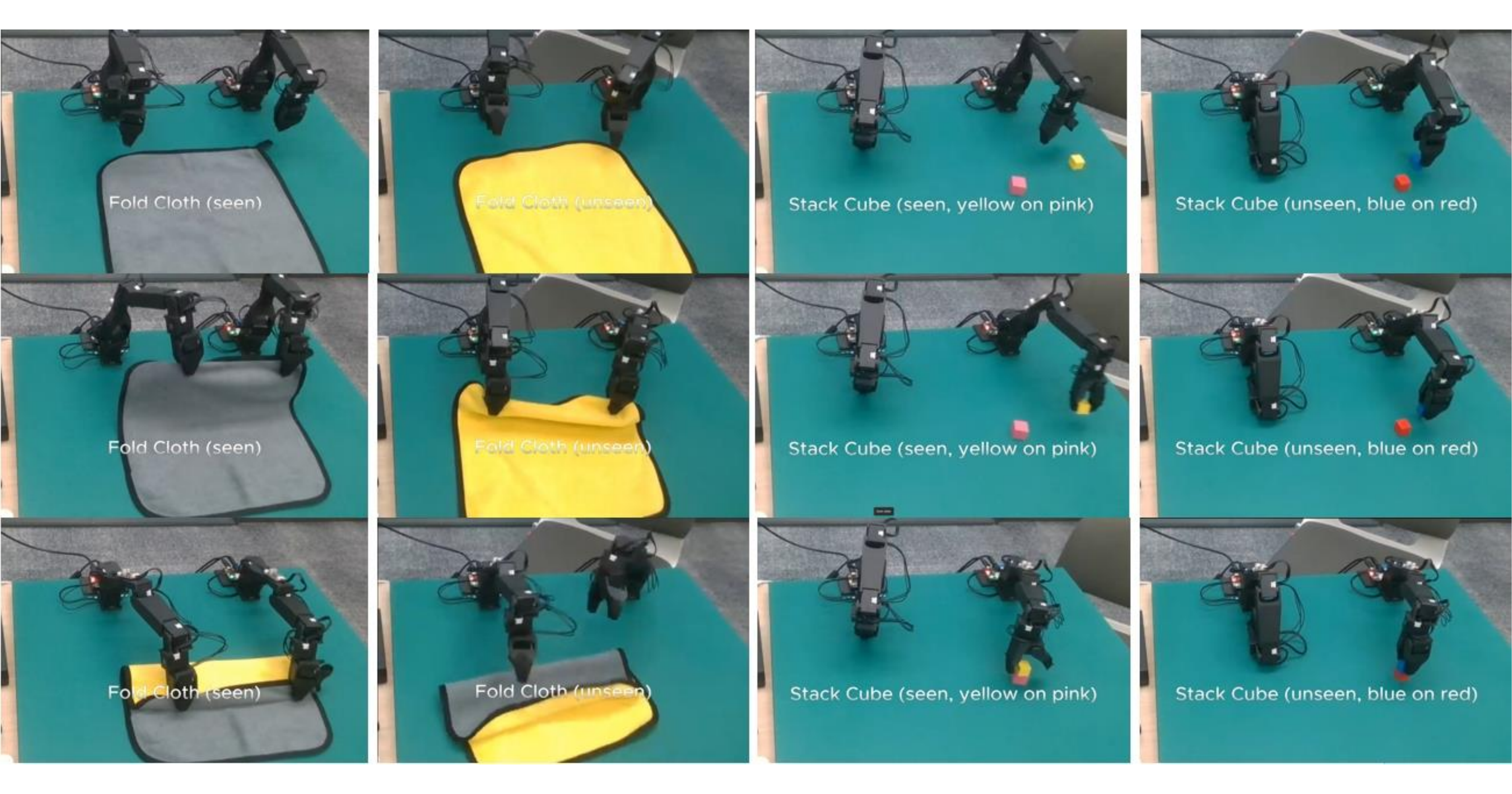} 
\label{method}
\caption{The demos of folding cloth and stacking cube skills.}
\label{pipelines}
\label{222}
\end{figure*}

The success rates for folding (seen), folding (unseen), stacking (seen), and stacking (unseen) are \textbf{85\%}, \textbf{65\%}, \textbf{50\%}, and \textbf{30\%}, respectively. 

} 


\vspace{-2mm}

\section{Discussion and Limitations}

Although our approach presents a promising direction for achieving generalizable robotic manipulation, there remain some clarifications and future works. 


\textbf{Which manipulation skills might benefit from our paradigm?} Our paradigm provides generalizable semantic, geometry, and temporal correlation priors in interacting objects. Therefore it can benefit all skills. As we observe that many end-to-end policy models fail to accurately locate objects due to lacking the equivariance of translation and rotation. Of course, we acknowledge that this requires manually designing complex prompt templates. We leave this issue to future works, such as better code generation. In addition, for contact-rich skills, extensive demonstrations are still needed to learn complex behaviors.

\textbf{How scalable is our paradigm?} Our paradigm aims to fully unify the generalizability of internet-scale models and the potential of imitation learning to capture multimodal action distribution from human skills at the lowest possible training and inference cost. We believe it is difficult for robot data to reach the scale of internet data, and even RT-1 cannot handle unseen object categories. Additionally, a large policy model such as RT-2 is difficult to interpret and incurs significant training and inference costs. Therefore, our multi-model paradigm has strong scalability, and improving the respective performance by scaling laws and coordination ability of various components in this system is a worthwhile exploration in the future. 


\textbf{How to improve the performance of our paradigm?} We observe that the main performance bottleneck of our system lies in the connection between the language reasoning module and the detection module, as current detectors still lack many visual concepts. Therefore, a good solution is to add prompts in the LLMs: \textit{Please add some descriptions of shape and color to the objects}. In addition, our system does not depend on completely precise masks and these situations are included in our training data.


\textbf{How to improve the execution speed of our system?} The integration of foundation models in our system introduces a more noticeable latency problem. To facilitate more efficient deployment in real-world scenarios, we also recommend offline language models and distilled lightweight vision models, such as MixformerV2~\cite{mixformerv2} and MobileSAM~\cite{mobile}. We report a comparison of the reasoning success rate of several language models in complex environments in Table~\ref{llm}. Although GPT-4 still performs the best, using a relatively lightweight dedicated model (DetGPT~\cite{detgpt}) is still a good choice. In addition, as for RTX A6000 GPU, we report the specific inference times of different modules and model sizes in Table~\ref{speed}, which indicates that distilled lightweight models have great value for use.


\vspace{-2mm}

\section{Conclusion}
In this paper, we propose to improve the generalization abilities of robotic manipulation by transferring internet-scale vision foundation models. By utilizing specified language-reasoning mask as our condition representation, which incorporates rich semantic, geometry, and temporal correlation priors derived from vision foundation models, into the policy model, our approach significantly improves the sample efficiency. In addition, our two-stream policy model also performs excellently with its local-global manner. Extensive experiments demonstrate the effectiveness and generalization capabilities of our paradigm and policy model, especially for unseen objects and backgrounds. Finally, we show that our policy model can be conditioned by assigning different values to object masks for different skills, therefore our system is scalable for new skills.

\section*{Acknowledgement}
This work is supported by the National Key R$\&$D Program of China (No. 2022ZD0160900), the National Natural Science Foundation of China (No. 62076119), the Fundamental Research Funds for the Central Universities (No. 020214380119), Jiangsu Frontier Technology Research and Development Program (No. BF2024076), and the Collaborative Innovation Center of Novel Software Technology and Industrialization.

{\small
\bibliographystyle{ieee_fullname}
\bibliography{egbib}
}

\end{document}